\documentclass[10pt,twocolumn,letterpaper]{article}

\usepackage{cvpr}
\usepackage{times}
\usepackage{epsfig}
\usepackage{graphicx}
\usepackage{amsmath}
\usepackage{amssymb}
\usepackage{soul,color}
\usepackage{amsmath}
\usepackage[ruled,lined]{algorithm2e}
\usepackage{tabularx}
\usepackage{booktabs}
\usepackage{epstopdf}
\usepackage{color, colortbl}
\usepackage{makecell}
\usepackage{multirow}
\usepackage{caption}
\usepackage[pagebackref=true,breaklinks=true,colorlinks,bookmarks=false]{hyperref}
\cvprfinalcopy
\def\cvprPaperID{8261} % *** Enter the CVPR Paper ID here

\begin{document}

%%%%%%%%% TITLE
\title{A Simple Fine-tuning Is All You Need:\\ Towards Robust Deep Learning Via Adversarial Fine-tuning}

\author{Ahmadreza Jeddi,
        Mohammad Javad Shafiee,
        Alexander Wong \\
% \textsuperscript{\rm 1}Computer Science Dept. University of Waterloo, Waterloo, Ontario, Canada\\
Waterloo AI Institute, University of Waterloo, Waterloo, Ontario, Canada\\

\{a2jeddi, mjshafiee, a28wong\}@uwaterloo.ca
% \\ a28wong@uwaterloo.ca
}
\date{}

\maketitle
\thispagestyle{plain}
\pagestyle{plain}

%%%%%%%%% ABSTRACT
\begin{abstract}
  \vspace{-2mm}
    Adversarial Training (AT) with Projected Gradient Descent (PGD) is an effective approach for improving the robustness of the deep neural networks. However, PGD AT has been shown to suffer from two main limitations: i) high computational cost, and ii) extreme overfitting during training that leads to  reduction in model generalization. While the effect of factors such as model capacity and scale of training data on adversarial robustness have been extensively studied, little attention has been paid to the effect of a very important parameter in every network optimization on adversarial robustness: the learning rate. In particular, we hypothesize that effective learning rate scheduling during adversarial training can significantly reduce the overfitting issue, to a degree where one does not even need to adversarially train a model from scratch but can instead simply adversarially fine-tune a pre-trained model.  Motivated by this hypothesis, we propose a simple yet very effective adversarial fine-tuning approach based on a `slow start, fast decay' learning rate scheduling strategy which not only significantly decreases computational cost required, but also greatly improves the accuracy and robustness of a deep neural network.  Experimental results show that the proposed adversarial fine-tuning approach outperforms the state-of-the-art methods on CIFAR-10, CIFAR-100 and ImageNet datasets in both test accuracy and the robustness, while reducing the computational cost by 8--10$\times$. Furthermore, a very important benefit of the proposed adversarial fine-tuning approach is that it enables the ability to improve the robustness of any pre-trained deep neural network without needing to train the model from scratch, which to the best of the authors' knowledge has not been previously demonstrated in research literature.
\end{abstract}

%%%%%%%%% BODY TEXT
\vspace{-0.35cm}
\section{Introduction}
The phenomenon of adversarial examples~\cite{szegedy2013intriguing} poses a threat to the deployment of the deep neural networks (DNNs) in safety and security sensitive domains~\cite{athalye2018obfuscated, kurakin2016adversarial, yuan2019adversarial}. Therefore, in the last few years, a huge body of research~\cite{goodfellow2014explaining, madry2017towards, papernot2016distillation, tramer2017ensemble, xie2019feature} has been conducted to improve the robustness of deep neural network models against the adversarial attacks. Another line of work in this field~\cite{goodfellow2014explaining, ilyas2019adversarial, ma2018characterizing, tanay2016boundary} focuses on understanding this phenomenon and the reasons why adversarial examples exist.

\begin{figure}[!]
\vspace{-0.6cm}
    \centering
    \setlength{\tabcolsep}{0.01cm} 
    \includegraphics[width=6cm]{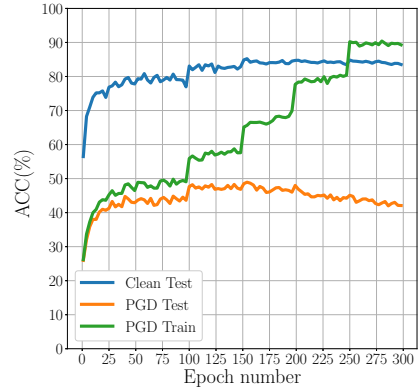}
    \vspace{-0.4cm}
    \caption{Accuracies of a PreAct ResNet18 classifier trained on CIFAR-10 dataset in 3 scenarios of a) clean test data, b) PGD attack on test data, and c) PGD attack on train data. While the robustness on the training data keeps improving through the training, on the test data for both the clean and PGD attack scenarios, the model reaches its maximum performance very early during the training and further training only reduces the model's generalization. This effect is worse on the test data robustness.}
    \label{fig:cifar_overfitting}
    \vspace{-0.55cm}
\end{figure}

Progress has been made in this area; especially, some recent works~\cite{cohen2019certified, jeddi2020learn2perturb, li2019certified, madry2017towards} have offered some levels of certified robustness against adversarial examples. However, there are still two main problems remaining for having robust deep neural network models; first, the level of the robustness is not very effective yet, and the second problem is that the current robust training algorithms are computationally very expensive with very high training times. This specially makes them impractical for the real-world problems with large sizes of training data.

% Many approaches have been proposed to train robust models~\cite{carmon2019unlabeled, guo2017countering, ross2017improving, xie2019feature, zhang2019defense}. However, a large number of these approaches fail to generalize to different adversarial attack scenarios. 
The simple AT approach remains the most popular and effective adversarial defense mechanism; especially, since Madry {\it et al.}~\cite{madry2017towards} introduced PGD adversarial attack and empirically illustrated that PGD is the universal first-order adversary (i.e. no other adversarial algorithm that uses first-order gradients can be more effective than PGD in fooling DNNs), AT with a PGD adversary has been the \textit{de facto} adversarial defense mechanism. This is mainly due to the robustness guarantee that PGD AT can provide, such that if a model is robust against PGD, then it is robust against all the other first-order adversaries as well. As a result of this certified robustness, almost all the recent state-of-the-art methods~\cite{carmon2019unlabeled, he2019parametric, zhang2019defense} take advantage of PGD AT as a part of their algorithms.  

PGD AT is an iterative and computationally expensive approach, which on average needs 8--10$\times$  more computational resources than a usual DNN training. As such, PGD AT lacks scalability and is not very practical for real-world problems. Even though some recent methods~\cite{shafahi2019adversarial, wong2020fast} have been able to improve the training time by applying complex modifications to the PGD AT method, they face a reduction in the robustness of the model as their limitation. In other words, these approaches face a trade-off between the training time and the robust generalization.

In this work, we demonstrate how by taking a different view at the AT approach, it is possible to reduce the training time and improve the scalability of AT by a large degree. Using the proposed algorithm  not only faces no loss on the accuracy and the robustness of the model, but it can also significantly improve the robust generalization of the model at the same time. Therefore, our adversarial fine-tuning approach mitigates the existing trade-off between the training time of AT and the model accuracy and robustness.

Motivated by the finding of Schmidt \textit{et al.}~\cite{schmidt2018adversarially} that during PGD AT a model highly overfits on the training data,  we hypothesize that this issue is partially related to learning rate scheduling at the training stage, and effective learning rate scheduling can mitigate the overfitting issue significantly. Figure~\ref{fig:cifar_overfitting} illustrates the overfitting problem during the training of a CIFAR-10 classifier. As seen, as the scheduler decreases the learning rate, the robustness on the training data gets better and better; however, not only does training for more epochs not improve the performance, but also causes a notable drop in the accuracy on the test data in  both the clean data (natural samples) accuracy and the test data robustness. As such, the model drastically overfits on the training data. Although Schmidt {\it et al.}~\cite{schmidt2018adversarially} defined the notion of sample complexity and  showed that in order to have a better robust generalization more data is needed, we argue that this  may not be very viable for real-world applications, especially, considering the very high computational overhead of PGD AT.

Experimental results show that by taking a different view on the PGD AT, the proposed approach is able to improve the robust generalization of the DNN models, achieving state-of-the-art performance on many  well-known datasets. Furthermore, one of the main benefits of the proposed adversarial fine-tuning algorithm is that it can be applied to any pre-trained model to increase its robustness.  This is specially important when dealing with AT of models with very large training data sizes (e.g. ImageNet) or in scenarios where a model has been trained using special techniques which may not be reproducible, such as when the model is trained by using a pipeline of transfer learning, or when weak or semi supervision is applied on billion scale datasets~\cite{yalniz2019billion}. In such scenarios, the usual PGD AT will not be practical due to its very high computational overhead, whereas adversarial fine-tuning not only is very computationally feasible and scalable, but it can also improve the adversarial robustness by decreasing the overfitting.

The main contributions of this work can be summarized as follows:
\begin{itemize}
    \item We introduce a simple yet effective strategy to visualize the embedding space of deep neural networks to help gain insights into why PGD AT results in overfitting and reduction in model generalization.
    \item We empirically explore the effect of learning rate on adversarial robustness of a deep neural network and demonstrate the importance of learning rate scheduling design on both convergence and generalization during adversarial training.
    \item We introduce a simple yet effective adversarial fine-tuning approach based on a `slow start, fast decay' learning rate scheduling strategy that can reduce computational cost of PGD AT by as much as $\sim$10$\times$, while at the same time noticeably improve the robustness and generalization of a deep neural network, and demonstrate its efficacy across three different datasets (CIFAR-10, CIFAR-100, and ImageNet) compared to state-of-the-art AT strategies. 
    \item We demonstrate for the first time, to the best of our knowledge, the ability to improve the robustness of any pre-trained deep neural network without the need to adversarially train a model from scratch. 
\end{itemize}

\section{Related work}
Followed by the seminal work of Szegedy {\it et al.}~\cite{szegedy2013intriguing} introducing the phenomenon of adversarial examples for DNN models, a huge body of research~\cite{carlini2017towards, goodfellow2014explaining, madry2017towards, stutz2019disentangling, tanay2016boundary} has been done on understanding this phenomenon and proposing solutions in order to overcome this weakness of deep neural network models. Among the many techniques, Adversarial Training (AT) ~\cite{goodfellow2014explaining, madry2017towards} has been the most popular and practical approach, mainly due to its simplicity of implementation, no additional inference cost, and most importantly its effectiveness. In AT, the adversarial examples are generated during the training and are used as the training samples; therefore, the most important component of the AT is its adversary (i.e., the algorithm that generates the adversarial samples). The effectiveness of the AT mainly depends on its choice of the adversary algorithm; as such, many AT algorithms do not offer much robustness against other adversaries, due to their lack of universally optimal optimization. However, Projected Gradient Descent (PGD) algorithm proposed by Madry {\it et al.}~\cite{madry2017towards} is the adversary that overcomes this issue.

PGD is an iterative algorithm which uses the first-order gradients of the loss function with respect to the input data to craft optimal perturbations to fool DNN models. Madry \textit{et al.}~\cite{madry2017towards} empirically showed that for a given input sample, no other adversary can find better perturbations than those of the PGD, hence they claimed that PGD is a universal first-order adversary. A very important implication of this claim is that if models are robust to PGD, they are robust against any other adversary, therefore, training a model by using PGD adversary (i.e., PGD AT) can yield robustness guarantees. As a result of the certified robustness, PGD adversarial robustness has been wildly popular in the literature~\cite{athalye2018obfuscated, kannan2018adversarial, shafahi2019adversarial, tsipras2018robustness, zhang2019you} and almost every successful technique in the recent years~\cite{carmon2019unlabeled,  hendrycks2019using, jeddi2020learn2perturb, liu2018adv, zhang2019defense, zhang2019theoretically} has had PGD AT as a part of its training pipeline. Learn2Perturn and PNI frameworks~\cite{jeddi2020learn2perturb, he2019parametric} combine PGD AT  and network randomization. AdvBNN~\cite{liu2018adv} adds PGD AT and Bayesian neural networks together, Zhang {\it et al.}~\cite{zhang2019defense}  generate adversarial samples by taking a group of samples into consideration instead of a single one. Camron {\it et al.}~\cite{carmon2019unlabeled} and Alayrac {\it et al.}~\cite{alayrac2019labels} augment the training data with unlabeled data and show how the results of PGD AT are improved.

Despite the relative effectiveness of the PGD AT, since this technique on average increases the training time of the models by 10 times, applying it on large-scale datasets might be very costly and challenging; especially, the recent works of the Camron {\it et al.}~\cite{carmon2019unlabeled} and Schmidt {\it et al.}~\cite{schmidt2018adversarially} showed that adversarial robustness requires more data, which means significant longer training time. In order to overcome this drawback of PGD AT, a group of techniques have been proposed to make the computational cost of the PGD AT more bearable. 

Instead of the usual PGD iterations, Free AT~\cite{shafahi2019adversarial} performs the usual gradient descent optimization on a batch of training data for \textit{m} times, while updating the adversarial perturbation of that batch by using the gradients of the loss function with respect to the input batch at the same time, and at each iteration perturbs the batch of input data with this perturbation. Therefore, they manage to take advantage of iterative AT, while reducing the training time as well. Fast AT~\cite{wong2020fast} uses Fast Gradient Sign Method (FGSM) adversary, which is the single iteration version of PGD, as its adversary, but, in order to have a PGD-like optimization, the injected perturbation is first randomly initialized and then updated by the gradients of the network. 

Aside from the complications that these methods add to the adversarial training, a major drawback of them is that they come at the cost of some drop in the accuracy and the robustness of the model. Therefore, the literature on this area has been facing a trade-off between improving the training time of the models and their robust generalization. On the other hand, our adversarial fine-tuning approach takes a different view of adversarial training and by mitigating the overfitting problem of PGD AT, not only improves the training time by 8--10$\times$ on  datasets such as ImageNet, CIFAR-10, and CIAFR-100, but it also significantly boosts the performance even going beyond the results of PGD AT itself.

% In order to overcome this issues we do this and we do that, we also have a set of results with increased data size (we can probably do this on CIFAR-10 if need be)

\section{Adversarial Training (AT)}
It has been shown that AT is an effective mechanism to improve the robustness of deep neural networks~\cite{goodfellow2014explaining, madry2017towards}. This approach was first introduced by Szegedy {\it et al.}~\cite{szegedy2013intriguing} where a mixture of adversarial and clean examples were used to regularize the deep neural networks and to help model learn how to cope with both types of natural and adversarial samples effectively. Goodfellow {\it et al.}~\cite{goodfellow2014explaining} extended this framework by generating the adversarial examples during the training by using an adversarial algorithm (i.e., FGSM in their work). The current adversarial training approaches still follow this setup, where adversarial examples are generated on-the-fly  during the training process. 

Madry {\it et al.}~\cite{madry2017towards} formulated AT as a min-max optimization problem:
\begin{align}
    \underset{\theta}{\min}\; \mathbb{E}_{(x,y)\in D}\Big[\underset{\delta \in S}{\max}\; L(\theta,x+\delta,y)  \Big]
    \label{eq:min-max}
\end{align}
where $\delta$ and \textit{S} show the perturbation and the boundary of perturbation, respectively. While the the inner optimization tries to find
the optimal adversarial perturbations, maximizing the loss, the outer minimization trains the model parameters $\theta$ such that the ``adversarial loss'', $L(\cdot)$, is minimized. 

Generally, any adversarial attack algorithm can be incorporated in the inner maximization part of the optimization. However, multi-step attacks and specially PGD are usually more powerful in providing effective perturbations. Especially, since Madry \textit{et al.} experimentally showed that PGD is the universal first-order adversary, this adversary has been wildly popular both for the adversarial training and for evaluating the ultimate robustness of deep models. An iterative PGD-k (PGD with k iterations) crafts the following adversarial example for a given natural sample \textit{x}:
\begin{align}
    x_{t+1} = \Pi_{x+S}\Big(x_t + \alpha\; sign (\nabla_x\;L(\theta,x,y)\Big)
\end{align}
where $\Pi(\cdot)$ is the projection function forcing the generated adversarial example remain within the boundary $S$ and $x_t$ is the adversarial example at step $t$, resulting from taking the ascent step of size $\alpha$.

\begin{figure}
\vspace{-0.5cm}
    % \centering
    % \begin{left}
    % \setlength{\tabcolsep}{0.1cm} 
    \begin{tabular}{cc}
             \includegraphics[width=4cm]{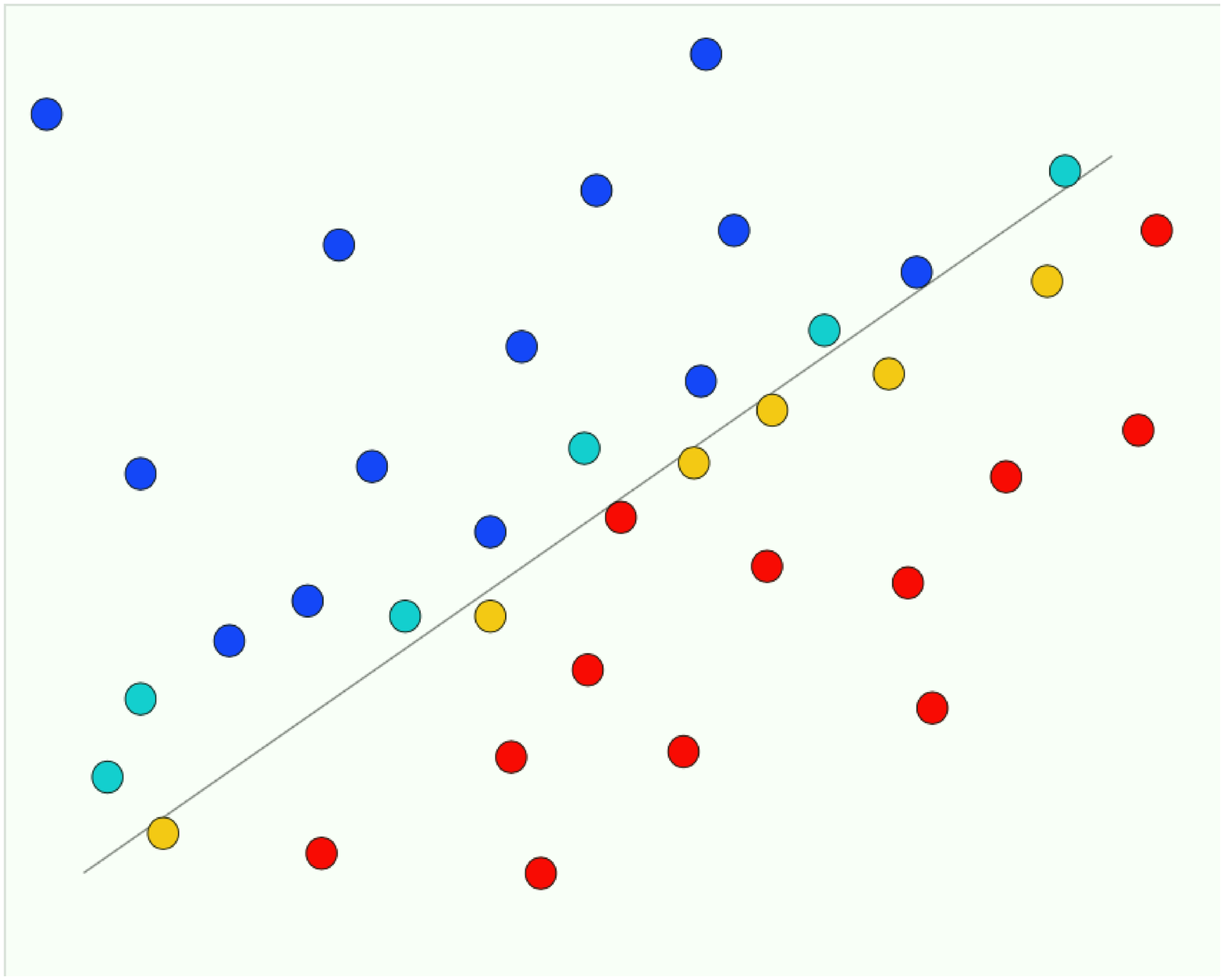}&   \includegraphics[width=4cm]{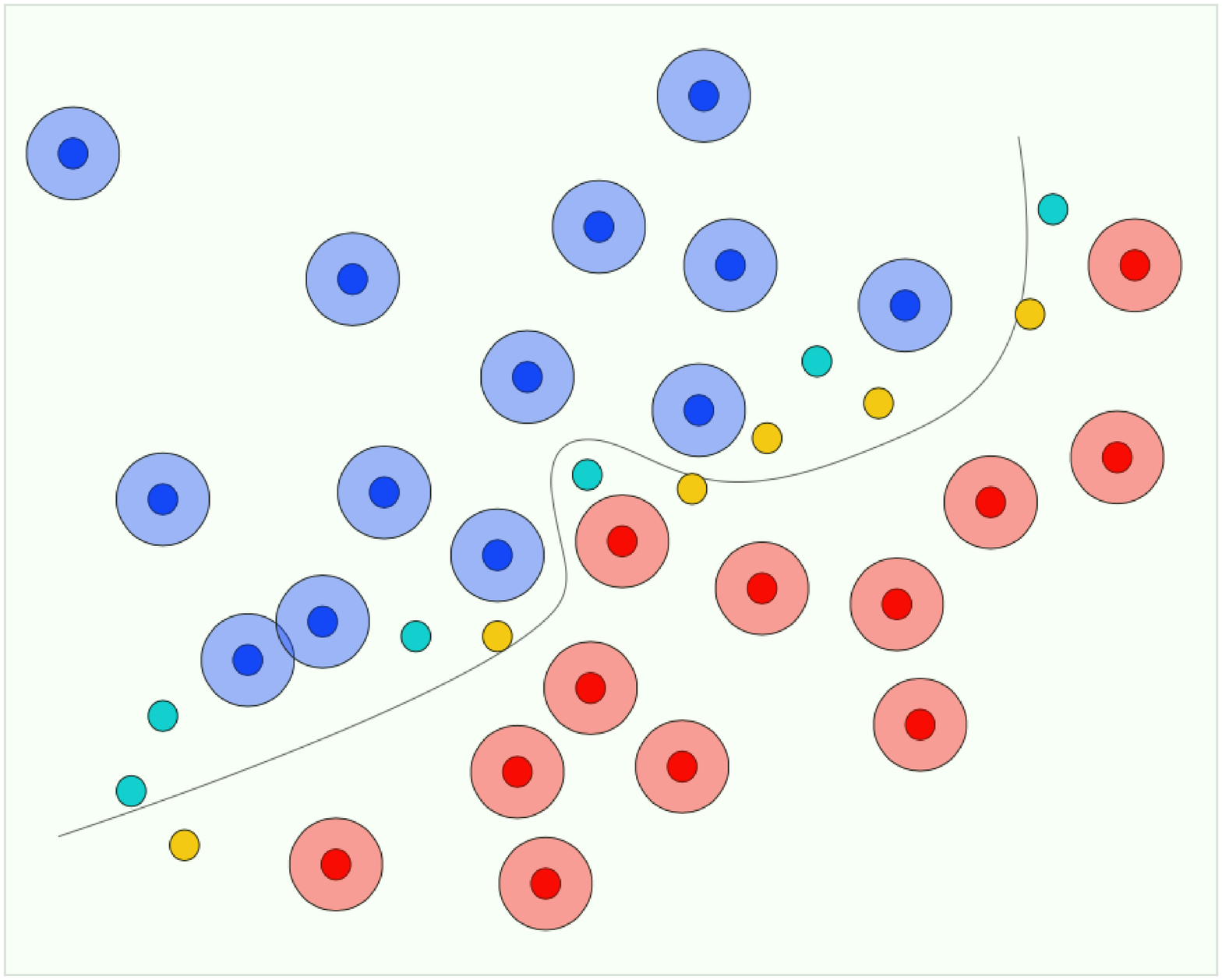} \\
         (a) & (b)
    \end{tabular}
    %   \end{left}
    \caption{Red: training samples of class one; Orange: test samples of class one; Blue: training samples of class two; Cyan: test samples of class two, and the bubbles show \mbox{$l_{\infty}$-ball}. (a) a model without PGD AT can fit a more simpler decision boundary and has a higher generalization on natural samples. (b) PGD AT helps the model to learn a decision boundary which in addition to assigning the label to the training samples, it assigns the same class label to $l_\infty$-ball around the sample as well. This causes the model to learn a more complex decision boundary and overfits on training data.}
    \label{fig:bubble_illustration}
    \vspace{-0.5cm}
\end{figure}

While increasing the number of steps $k$ would result in more powerful adversarial examples with higher loss, one important limitation is the high computational overhead of PGD. Computing the model's gradient for the input data in each step is the main bottleneck of this approach and as such increasing the number of steps $k$, increases the training time significantly. Recently, several adversarial defense techniques have attempted to speed up this process~\cite{shafahi2019adversarial, wong2020fast, zhang2019you}. These methods usually focus on reducing the frequency or scale of the required back-propagated gradients, so that the overall computational overhead decreases. However, most of these methods face a trade-off between the robustness and the training time of the model.

Intuitively speaking, the PGD AT approach tries to train the model to associate not only a single location in the space to the corresponding label of the sample $x$, but to associate a $l_{\infty}$-ball around the example $x$ to the same class label. Figure~\ref{fig:bubble_illustration} illustrates this phenomenon graphically; as seen, the resulted decision boundary is  more complex, so that the model is able to associate the same class label to the sample and $l_{\infty}$-ball around it.

While this approach helps the model to achieve some level of certified robustness, it suffers from two main limitations. I) It enforces a very high computational burden (as it is an iterative algorithm). PGD AT with its default setup is on average $\sim $8--10$\times$  more computationally complex than a model being trained only on natural samples.  II)  PGD AT highly overfits on the training data  resulting in drops in both the generalization of the model on natural samples and even the model robustness on test data. Figure~\ref{fig:bubble_illustration}(b) demonstrates this effect graphically; as the model struggles to learn how to assign the same label to the sample and $l_{\infty}$-ball around it, it looses its generalization specially near the decision boundary. As such, there is a high chance that the unseen data lying close to decision boundary are misclassified because of this overfiting issue. 
In the following section, we shed more light on the severe overfitting issue of PGD AT, and then, present our method to overcome the first limitation of PGD AT and largely mitigate the second limitation.

\subsection{Overfitting Issue}
\label{sec:overfit}
The overfitting issue associated with the PGD AT was first introduced by Schmidt {\it et al.}~\cite{schmidt2018adversarially}. They experimentally illustrated that PGD AT on the CIFAR-10 dataset can result in more than 50\% difference in the adversarial robustness of the model on the training and the test datasets. Figure~\ref{fig:cifar_overfitting} further validates the overfitting observation on the CIFAR-10 dataset. As seen,  the adversarial robustness of the training and test data diverge (i.e. overfitting on the training data) when the learning rate of the gradient descent optimizer is reduced; while the learning rate scheduling significantly boosts the robustness of the model on the training data,  a slight increase is followed by a smooth robustness decrease on the test data. This overfitting is a result of PGD AT trying to learn a whole $l_{\infty}$-ball instead of just a sample point in the space. This effect is illustrated in Figure~\ref{fig:bubble_illustration}, where in order to learn the whole bubbles around each sample, the model highly overfits on the training data. In this situation, training for more number of iterations only attributes to model learning the bubbles better and get worse generalization (Figure~\ref{fig:cifar_overfitting}). This overfitting trend is consistent  for other datasets such as CIFAR-100 and ImageNet as well.

It has been empirically shown~\cite{jeddi2020learn2perturb, madry2017towards, sun2019towards}, that increasing the number of training samples as well as larger network capacity can attribute to a better robust generalization. Especially, on a simulated Gaussian dataset with 2 classes, Schmidt {\it et al.}~\cite{schmidt2018adversarially} theoretically demonstrated that robust generalization requires higher \textbf{sample complexity}. The sample complexity refers to the number of required samples to secure a guaranteed robustness level. Moreover, they experimentally validated this effect on more complex datasets such as CIFAR-10 and SVHN, where increasing the training data size improves the adversarially robustness generalization. Camron \textit{et al.}~\cite{carmon2019unlabeled} further extended the sample complexity theory, and demonstrated that even the augmentation of the training data by applying supervision on unlabeled data can enhance the adversarial robustness.

\begin{figure}[!]
\vspace{-0.5cm}
    \centering
    \setlength{\tabcolsep}{0.01cm} 
    \includegraphics[width=6cm]{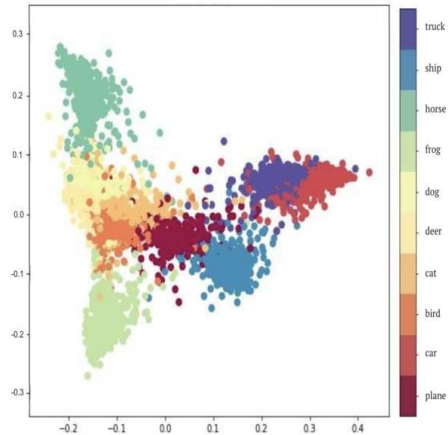}
    \vspace{-0.4cm}
    \caption{2D visualization of the embedding space of a WRN-28-10 classifier trained on CIFAR-10 dataset. Two feature reduction techniques (LDA and PCA) are combined to achieve this space which shows the relative location of the test data samples in the embedding space. While the meaningful semantic relations among class labels and their heterogeneous distributions further validate the requirement of sample complexity toward the robust generalization, other non-trivial factors   can  affect  the  sample  complexity  such  as  the inter-  and  intra- class  relations  among  the  data  samples. }
    \label{fig:cifar_space}
    \vspace{-0.7cm}
\end{figure}

Here, we further analyze the effect of sample complexity by visualizing the samples in the embedding space.  In order to have a better understanding of the latent sample space resulting from the model's embedding,  we take advantage of a simple yet very effective neural network manifold visualization technique. Our visualization method consists of two consecutive simple feature reduction steps; I) the supervised LDA method is used to reduce the feature space size to $c-1$ ($c$ is the number of classes), II) then an unsupervised PCA feature reduction is performed to reduce the space size to 2, which is suitable for visualization. Figure~\ref{fig:cifar_space} illustrates the result of applying this technique on the embedding space of a CIFAR-10 classifier. As seen, an intuitive semantic relation exists between different class labels in the embedding space; for example, the embedding associated to  different animal classes are close to each other, and the embeddings corresponding to vehicles are located in close space as well. At the same time, classes that have less semantic similarity to each other hold a further distance. 

One very important insight from this visualization is that, although increasing the number of samples can help improve the robust generalization, there are many non-trivial factors that can affect the sample complexity. For example the inter- and intra-class relations among the data samples, and the scale of the perturbations ($\epsilon$) may drastically increase the sample complexity in real applications; however, gathering such scales of data and training a model using PGD AT on them seems almost impossible. Therefore, achieving a desired robustness level via increasing the data size would be very impractical. In order to boost the model generalization despite the lack of sufficient data, we focus on the learning rate scheduling component of a model. We hypothesize that a smart learning rate scheduling during the training can significantly mitigate the overfitting issue. In the next section, we will provide empirical evidence to support this hypothesis.

% Therefore, we need to come up with smarter ways to overcome the overfitting issue; in the next section, we describe how we can improve this problem by using a smart learning rate scheduler. After describing the effect of the learning rate scheduling, we will propose our adversarial fine-tuning approach.

% \hl{we need one more paragraph here, explaining more about this phenomena, it would be in one hand ....., so then the next paragraph will be on the other hand...}

% On the other hand, xxx {\it et al.}~\cite{} attributed this issue to the \textbf{sample complexity} of the training data and experimentally showed that in order to achieve adversarially robust generalization, more training data should be provided during the training. Followed by that xxx {\it et al.}~\cite{} further illustrated that  the sample complexity theory can help to improve the robustness of the model and demonstrated that even the augmentation of the training data by unlabeled data can enhance the adversarial robustness. To do so,  semi-supervised techniques are applied to extract unlabeled data that have similar semantics to available annotated training example with ground truth. 

% \subsection{Sample Complexity}
% Take a look at the Madry paper and read more about sample complexity. According to our results, our method works better for CIFAR-100 than for CIFAR-10, could it be becuase of the number of samples per class and the overall space complexity? (the same pattern apparently exists for ImageNet as well)

% In order to achieve that, one needs a bigger model and more number of iterations, so even if we get the bigger model.
\vspace{-0.3cm}
\section{The Role of Learning Rate Scheduling}
 In this section, we analyze the relation between the model overfitting and learning rate scheduling. We show how the overfitting happens from the viewpoint of the learning rate scheduling; then, we demonstrate how by utilizing a smart learning rate scheduler it is possible to reduce the effect of overfitting for a given model and dataset and consequently improve the adversarial robustness and generalization of the model.

\begin{figure*}
\vspace{-0.85cm}
    \centering
    \begin{tabular}{ccc}
         \includegraphics[width=5cm]{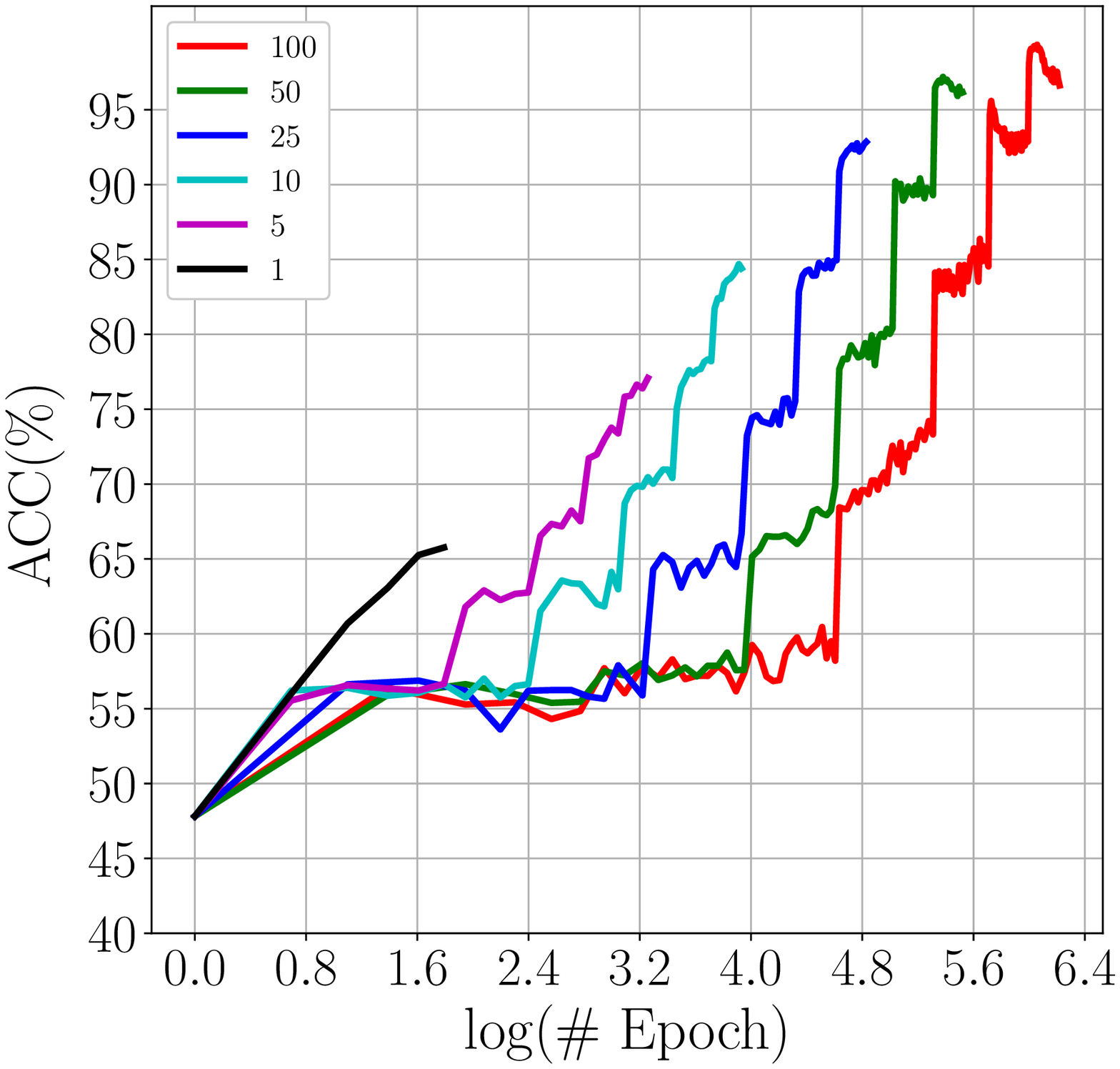} &\includegraphics[width=5cm]{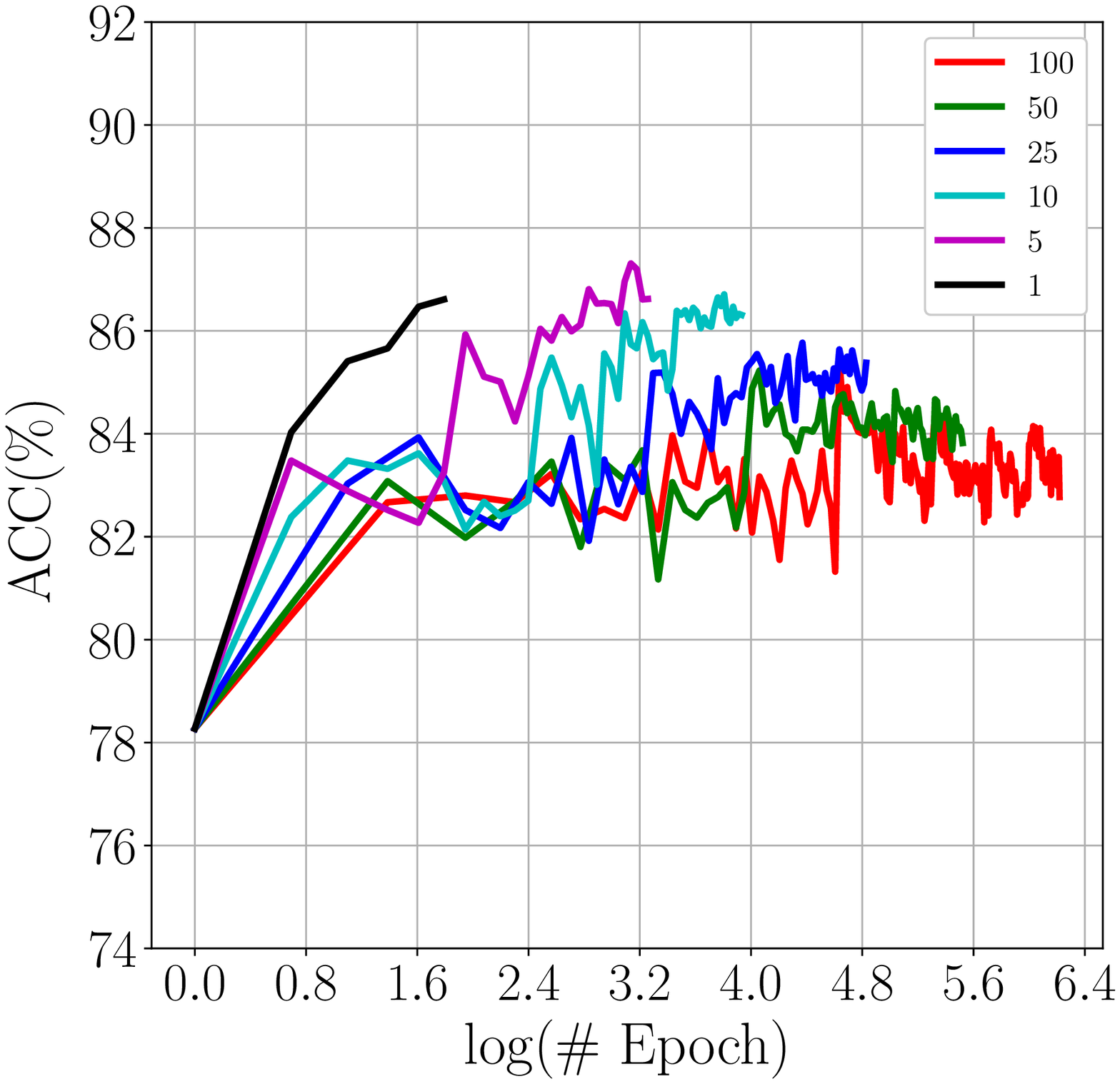} & \includegraphics[width=5cm]{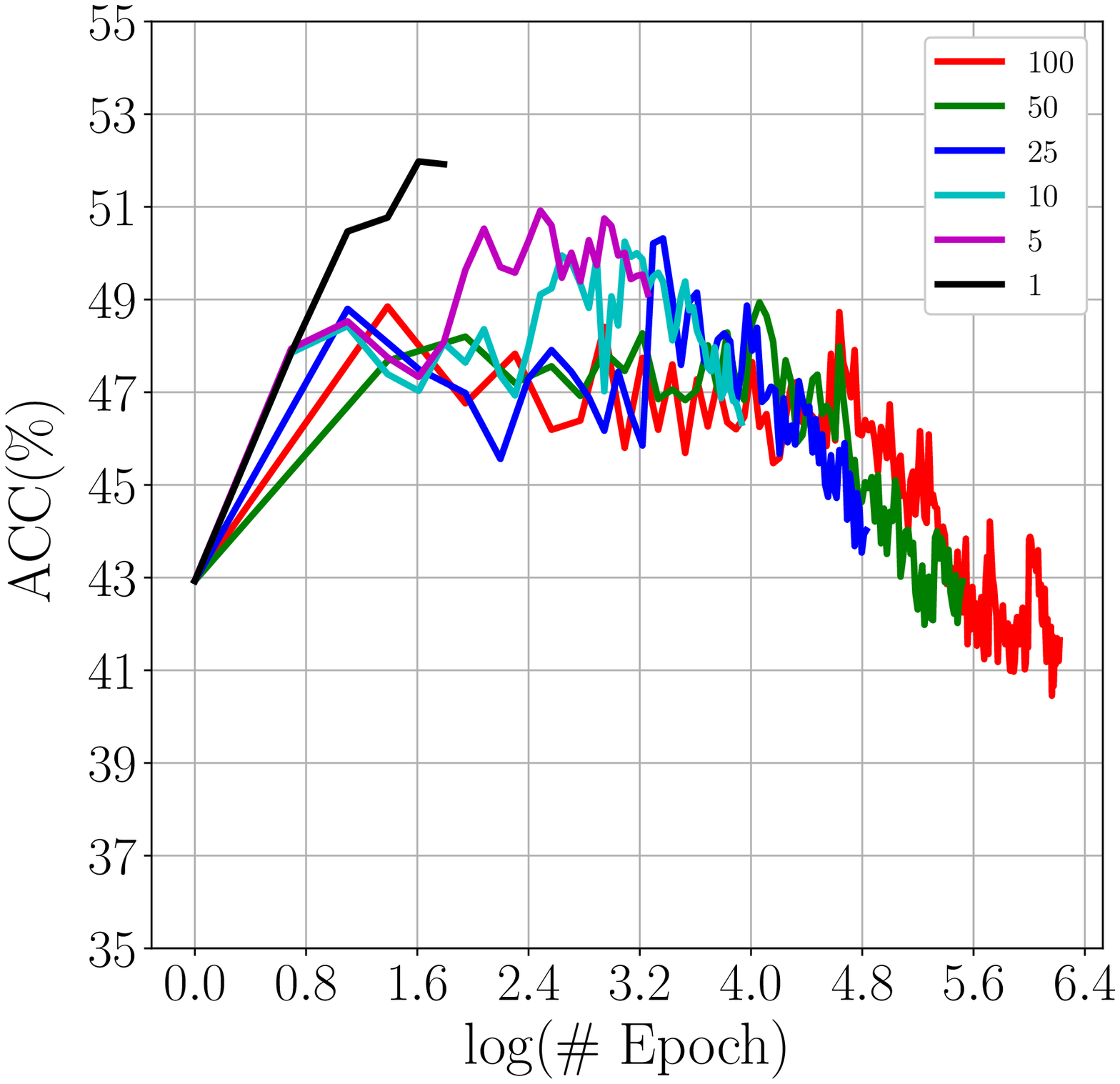} \\
         (a) Performance on Training Data & (b) Performance on Clean Test Data &(c) Performance on Adversarial Test Data
    \end{tabular}
    \vspace{-0.2cm}
    \caption{The effect of learning rate scheduling on model generalization and robustness; PGD AT with more number of epochs improves the model's robustness on the training data. However, bigger number of training iterations causes some drops in the model's accuracy and robustness on test data as evident in (b) and (c). As seen, training a model with less number of epochs can result in better reducing the overfitting issue and improves the adversarially robust generalization. }
    \label{fig:lr_scheduling}
    \vspace{-0.55cm}
\end{figure*}

Figure~\ref{fig:bubble_illustration} shows a binary classification problem  with examples from each class label which are overlaid with  $l_{\infty}-\epsilon$ neighborhood  shown by a disk around each sample. PGD AT tries to not only find boundaries that would correctly classify the examples, but it also tries to fit a model which  assigns the same label to  the whole $l_{\infty}-\epsilon$ neighborhood around the example. Since for the real image datasets such as CIFAR-10, CIFAR-100, and ImageNet, the amount of the available data fails to meet the sample complexity requirements of the robust generalization (i.e, consider the actual very high-dimensional space of these image sets) learning the whole bubble for the training data causes overfitting.

A very intriguing pattern that we observe in our experiments is that long plateaus in the learning rate scheduling of the model during training further contributes to the overfitting problem. For a gradient-based optimizer, we consider the learning rate scheduler formulated as \textit{step-LR}$(i, \gamma)$, where $i$ and $\gamma$ show the number of epochs for each plateau and the step scale, respectively. Figure~\ref{fig:lr_scheduler} compares the effect of using 6 different learning rate schedulers for fine-tuning a pre-trained  PreAct ResNet18 on CIFAR-10 dataset. The only difference between different runs in the Figure is the size of the plateau, and the step scale is 0.5 for all, so, all of these learning rate schedulers can be formulated as \textit{step-LR}$(i, 0.5)$. For the sake of a fair comparison, all trials use the exact same pre-trained model as their initialization. As seen in Figure~\ref{fig:lr_scheduler}, as $i$ increases, the model gets more time to thoroughly learn the bubble around each sample and therefore, the overfitting on the training data increases, resulting in a drop in both the accuracy and the robustness of model on the test data. On the other hand, for smaller values of $i$ the exact opposite trend happens which causes a decrease in the train data robustness and an increase in both the accuracy and the robustness of model on the test data, which means less overfitting. 

Motivated by the illustrated experiment and our observations regarding the sample complexity and the learning rate scheduling, we hypothesize that a simple adversarial \textit{fine-tuning} approach can mitigate the overfitting issue, and  achieve great robustness generalization.  It is worth mentioning that, although other factors such as the model capacity have important effects on the robust generalization of the trained models, in this work,  we only study the effects of the learning rate scheduling and the sample complexity of the training data on the model's robustness.

\subsection{Adversarial Fine-tuning}
Motivated by the empirical evidence on the significant impact of learning rate scheduling on adversarial robustness, we propose a simple yet effective adversarial fine-tuning (AFT) technique for not only reducing training time (and hence computational cost) but also improving the robustness of a deep neural network. More specifically, the proposed AFT approach comprises of two main aspects: 
\begin{itemize}
    \item \textbf{Model pre-training}: A model is trained regularly using natural samples without consideration of adversarial perturbations for stronger initial generalization.
    \item \textbf{`Slow Start, Fast Decay' fine-tuning}: The pre-trained model is fine-tuned using adversarial perturbations following a `slow start, fast decay' learning rate schedule for a small number of epochs for stronger adversarial robustness while preserving generalization.
\end{itemize}

This proposed technique is contrary to previously proposed AT methods that involve training models with adversarial perturbations in an end-to-end manner from scratch, which is significantly more computationally costly and lead to reduced model generalization.   Details of the two main aspects of the proposed AFT technique are described below.
\vspace{-0.35cm}
\subsubsection{Step 1: Model Pre-training}
The first step of the proposed AFT strategy involves pre-training a model regularly with natural samples.  Our experiments suggest that having a good pre-trained model is of high value, and we empirically find that the more data the pre-trained model is exposed to during its training, the better initialization it would be for the fine-tuning step. Experimental results validate this hypothesis on the CIFAR-10 and ImageNet datasets. This observation is particularly exciting since one can take advantage of already pre-trained models that have been trained on a very large set of data. This is especially important in many classification problems that leverage semi- or weak-supervision techniques to enrich their training data where an additional set of samples are used to improve the classification performance. SWSL~\cite{yalniz2019billion} is an example of such approach where billion-sample scale data~\cite{mahajan2018exploring, thomee2016yfcc100m} is used to achieve state-of-the-art performance on the ImageNet dataset. In our experiments, we show that adversarially fine-tuning such pre-trained models only on the main training data can improve the robustness and test accuracy by 7-8\%. It is worth mentioning that due to the high computational overhead of PGD AT, conducting PGD AT on the dataset augmented by weak or semi-supervised method is not feasible.  As such, we are motivated to introduce a `slow start, fast decay' finetuning strategy.
\vspace{-0.3cm}
\subsubsection{Step 2: `Slow Start, Fast Decay' Fine-tuning}
The second step of the proposed AFT strategy involves fine-tuning the pre-trained model using adversarial perturbations via a 'slow start, fast decay' learning rate schedule. Given the overfitting issue explained before and the tendency of neural network to catastrophically forget their previously learned distributions when exposed to new samples, it is very crucial that the selected learning rate scheduling helps the model learn the new distribution of adversarial samples without sacrificing the previously learnt knowledge (natural data examples). Therefore, it is important that the learning rate is slow at first, so that the model gradually learns the new distribution.

\begin{figure}
\vspace{-0.75cm}
    \centering
    \setlength{\tabcolsep}{0.01cm} 
    \includegraphics[width=7cm]{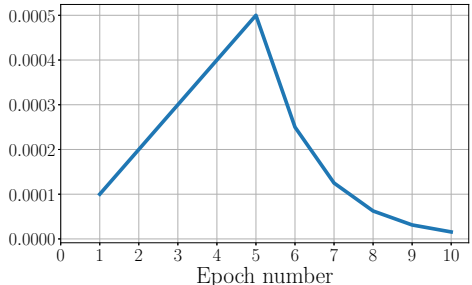}
    \vspace{-0.3cm}
    \caption{The 'slow start, fast decay' learning rate schedule introduced in the proposed adversarial fine-tuning strategy.}
    \label{fig:lr_scheduler}
    \vspace{-0.5cm}
\end{figure}

 The proposed `slow start, fast decay' learning rate scheduling strategy is shown in Figure~\ref{fig:lr_scheduler}. As seen in this Figure, we take advantage of a slow start learning rate scheduling, where the starting learning rate is chosen small, and then the learning rate is smoothly increased, so that the new distribution is learnt with a faster pace. This approach helps the model learn the distribution of the adversarial examples without forgetting the distribution of the natural samples. After these first few epochs, the learning rate is reduced very fast so that model performance converges to a steady state, without having too much time to overfit on the training data. 

\section{Experimental Results \& Discussion}
In this section, we evaluate the proposed adversarial fine-tuning (AFT) method on three well-known classification datasets of CIFAR-10, CIFAR-100, and ImageNet. Moreover, we compare the results with state-of-the-arts techniques and show how the proposed AFT algorithm outperforms  other method by  large margins. 

\subsection{Setup}
For CIFAR-10 and CIFAR-100 datasets~\cite{krizhevsky2009learning} we choose a wide residual network WRN-28-10~\cite{zagoruyko2016wide}. The pre-training on both of these datasets is done using an SGD optimizer (with momentum of 0.9 and weight decay of $5\times10^{-4}$) and they are trained for 200 epochs, with an initial learning rate of $0.1$ and the learning rate is multiplied by $0.2$ at the epochs 60,120, and 160. The dropout rate of 0.3 is used for the wide residual network training. For reporting results on ImageNet dataset, the pre-trained ResNet50~\cite{he2016deep} is used which is publicly available. 

For adversarial fine-tuning of all models  an Adam optimizer~\cite{kingma2014adam} with the same learning rate scheduling as shown in Figure~\ref{fig:lr_scheduler} is used and the models are  fine-tuned  for only 10 epochs. Moreover, the robustness of all the models is evaluated via  PGD adversary with 20 iterations.

In order to demonstrate the effect of having a good network initialization in our fine-tuning approach, we take advantage of two  recently proposed semi-supervised techniques which improved the performance of models trained on CIFAR-10 and ImageNet.  A pre-trained ResNet50 model which trained via the Semi\&Weakly Supervised Learning (SWSL) proposed by Yalniz \textit{et al.}~\cite{yalniz2019billion} is used to evaluate the proposed method on ImageNet dataset. SWSL trains ImageNet models by using a semi-supervision method on billion scale data. We refer to this model as ResNet50-SWSL. For reporting the result on CIFAR-10 dataset, we take advantage of the training data augmentation technique introduced by Carmon \textit{et al.}~\cite{carmon2019unlabeled}, where they apply semi-supervision on the 80 Million Tiny Images dataset~\cite{torralba200880} to augment the CIFAR-10 dataset with 500K additional data; in our experiments we pre-train the WRN-28-10 on CIFAR-10 augmented with this 500K unlabeled data by following the same approach the authors proposed and we refer to this model as CIFAR-10+500K. It is worth noting that due to the large volume of these semi-supervised data augmentations, using them in the PGD AT is highly impractical, therefore, we fine-tune the models only on the original training datasets.

\begin{table}
\vspace{-0.55 cm}
\caption{Evaluation results on CIFAR-10 dataset; the proposed algorithm is compared with  the state-of-the-art methods which have been proposed in the recent years to improve the efficiency and the performance of (PGD) AT. The competing methods aim to provide an efficient approach in AT while reducing the computational complexity compared to original PGD AT (PGD AT). As seen, the proposed fine-tuning algorithm can result to higher accuracy on clean data while outperforms others significantly in robustness against PGD attack. Result (AFT (+500K)) shows that a model with better initialization can offer higher robustness after performing adversarial fine-tuning algorithm. }
\vspace{-0.03cm}
\centering
 \small
\setlength{\tabcolsep}{0.1cm} 
\begin{tabular}{l|lccc}
       \bf Method & \bf Architecture & \bf Clean & \bf PGD   & \bf Time (min) \\
         \hline
        \bf Natural & \small WideRes-32x10 & 95.01 & 00.0 & 780
       \\ 
        \bf PGD AT~\cite{madry2017towards} &\small WideRes-32x10 & 87.25 & 45.84 & 5418
       \\ 
       \bf  Free AT~\cite{shafahi2019adversarial} &\small WideRes-32x10 & 85.96 & 46.82 & 785
       \\ 
        \bf Fast AT~\cite{wong2020fast} &\small PreAct ResNet18 &  83.81 & 46.06 & 12 
       \\
        \bf YOPO~\cite{zhang2019you} &\small WideRes-34x10 &  86.70 & 47.98 & 476 
       \\
        \bf ATTA~\cite{zheng2020efficient} &\small WideRes-34x10 &  85.71 & 50.96 & 134
       \\
       \hline
        \bf AFT &\small WideRes-28x10 & \bf 88.15 & \bf 51.7 & 486 
       \\
        \bf AFT (+500K) &\small WideRes-28x10 & \bf 88.42 & \bf 52.8 &  486 
       \label{tab:cifar10}
    %   \vspace{0.7cm}
\end{tabular}
\vspace{-0.5 cm}
\end{table}

\subsection{Competing Methods}
We compare our proposed method with the following state-of-the-art approaches which utilize PGD AT for improving adversarial robustness.\\
% \begin{enumerate}
    \textbf{Free AT}~\cite{shafahi2019adversarial}: Instead of using the regular PGD AT, they do the FGSM AT on the same batch for \textit{m} times, while updating the gradients of the input in each iteration.\\
    \textbf{Fast is better than free (FAST AT)}~\cite{wong2020fast}: They use FGSM instead of PGD, but in order to get PGD-like optimization power they initialize the gradients randomly within the $l_{\infty}$ ball.\\
    \textbf{You Only Propagate Once (YOPO)}~\cite{zhang2019you}: They show that PGD updates are coupled with the first layer of DNN, so they restrict the adversary updates to the first layer, hence, reducing the computational cost.\\
    \textbf{Efficient AT with Transferable Adversarial Examples (ATTA)}~\cite{zheng2020efficient}: They find that adversarial examples from previous training epochs still remain adversarial in the next epochs as well, and propose a framework which utilizes this transferability effect.
% \end{enumerate}
\vspace{-0.2cm}
\subsection{Results}
\vspace{-0.2cm}
As the first experiment, the proposed method and the competing algorithms are compared via CIFAR-10 dataset, and the robustness of the model are evaluated against a PGD adversary with $\epsilon = \frac{8}{255}$. As seen in Table~\ref{tab:cifar10}, the proposed fine-tuning algorithm can provide models with both highest  generalization on natural images (accuracy on clean data) and greatest robustness against adversarial attack. Results show that using data augmentation and taking advantage of 500K additional data samples  to augment the CIFAR-10 dataset improves the robustness of the model against adversarial attack significantly as well.
It is important to note that this additional data samples are not used in adversarial fine-tuning step but only in the training of the model on natural images. As such, the result confirms the hypothesis that a model with a higher generalization can offer better robustness against adversarial attacks when trained properly.

\begin{table}
\vspace{-0.55 cm}
\caption{CIFAR-100 experimental results; the accuracy and PGD robustness of the proposed method and  the state-of-the-art methods are compared against PGD adversarial attack. Two different  $\epsilon $ (AFT ($\epsilon = \frac{8}{255}$) and AFT ($\epsilon = \frac{10}{255}$)) are used in the proposed fine-tuning technique to illustrate the effect of PGD adversarial training in model robust generalization.}
\vspace{-0.2cm}
\centering
\small
\setlength{\tabcolsep}{0.1cm} 
\begin{tabular}{l|cccc}
       \bf Method & \bf Architecture &
       \bf Clean & \bf PGD   & \bf Time (min)\\
          \hline
       \bf Natural & WideRes-32x10 & 80 & 00.00 & 817
       \\ 
       \bf Natural & WideRes-28x10 & 82 & 00.00 & $\sim$ 750
       \\
       \bf PGD AT~\cite{madry2017towards}  & WideRes-32x10 & 60 & 22.50 & 5157
       \\ 
       \bf PGD AT~\cite{madry2017towards}  & WideRes-28x10 & 62 & 20.50 & $\sim$ 5000
       \\ 
       \bf Free AT~\cite{shafahi2019adversarial} & WideRes-32x10 & 62.13 & 25.88 & 780
       \\  
       \hline
       \bf AFT ($\epsilon = \frac{8}{255}$) & WideRes-28x10 &  \bf 68.15 & 23.29 & 486 \\
       \bf AFT ($\epsilon = \frac{10}{255}$) & WideRes-28x10 &  66.57 & \bf 25.12 & 486
       \label{tab:cifar100}
    %   \vspace{0.7cm}
\end{tabular}
\vspace{-0.7cm}
\end{table}

To confirm the effectiveness of the proposed algorithm as the second experiment, it is evaluated via CIFAR-100 dataset. A same setup as CIFAR-10 experiment is used, where a PGD adversary with 20 iterations and $\epsilon = \frac{8}{255}$ is utilized to evaluate the robustness of the competing methods. Results reported in Table~\ref{tab:cifar100} further illustrates the effectiveness of the proposed algorithm in providing robust DNN models while does not sacrifice the model's generalization on neutral images. To better analyze the effect of PDG adversarial training in the proposed fine-tuning technique, two different $\epsilon$ values (AFT ($\epsilon = \frac{8}{255}$) and AFT ($\epsilon = \frac{10}{255}$)) have been used to trained the model. As seen, while using perturbed images with stronger attack can improve the robustness of the model, it resulted a drop in the accuracy of the model against natural data samples which further validates the overfitting issue explained in Section~\ref{sec:overfit}. Higher value of $\epsilon$ means bigger $l_\infty$-ball around the samples and this forces the model to use more complex decision boundary to fit on the data. 
The proposed fine-tuning techniques is more than $10 \times$ faster than the conventional PGD adversarial training method (PGD AT) and is it even $\sim$2$\times$ faster compared to Free AT algorithm in training the final robust model. 

% \subsubsection{Robustness on CIFAR10}
% We compare the robustness of our model and the competing approaches against a PGD adversary with $\epsilon = \frac{8}{255}$. The results are shown in Table~\ref{tab:cifar10}.

As the last experiment, the proposed algorithm and the competing methods are compared against ImageNet dataset. To evaluate the model  $\epsilon = \frac{2}{255}$ is chosen for the PGD adversarial attack. As seen in Table~\ref{tab:ImageNet}, while the proposed fine-tuning technique outperforms the competing methods in clean accuracy which shows the generalization of the DNN model on natural images, it provides comparable robustness against adversarial attack. This is evident by the reported result for ResNet50 network architecture.
The reported result for ResNet50-SWSL  demonstrates the significant effect of pre-training  and the effect of the model generalization on the robustness result. The  ResNet50-SWSL  architecture is further trained via a semi-supervised technique as mentioned in Setup Section.
As seen, this further training can result a significant boost in both the generalization of the model and model accuracy on clean data and robustness of the model against adversarial attack. Results show that the robustness of the model can improve by more than 7\% and outperforms competing methods significantly while it can provide the final model in reasonable time-frame.

% \subsubsection{Robustness on CIFAR100}
% For the CIFAR100 we follow the same setup for the evaluation as CIFAR10, where a PGD adversary with 20 iterations and $\epsilon = \frac{8}{255}$ is utilized to evaluate the robustness of the competing methods.

% \subsubsection{Robustness on ImageNet}
% The robustness evaluations on ImageNet classification are done with $\epsilon = \frac{2}{255}$. You can see the effect of semi-supervision and ....

\begin{table}
\vspace{-0.5 cm}
\caption{Comparison results on ImageNet dataset; The proposed method outperforms competing algorithms on clean data samples (natural images) while provide comparable robustness performance as evident by ResNet50 results. The reported result for ResNet50-SWSL architecture shows the significant effect of pre-training  and the generalization of the model on robustness. As seen, the model offers $\sim7\%$ robustness improvement. }
\vspace{-0.1cm}
\centering
\small
\setlength{\tabcolsep}{0.1cm} 
\begin{tabular}{l|lccc}
       \bf Method & \bf Architecture &
       \bf Clean & \bf PGD   & \bf Time (hours)\\
          \hline
          
       \bf Natural & ResNet50 & 76.04 & 0.13 & -
       \\ 
       \bf PGD AT~\cite{madry2017towards}  & ResNet50 & 68.0 & 45.0 & $\sim$280
       \\ 
       \bf Free AT~\cite{shafahi2019adversarial} & ResNet50 & 64.5 & 43.5 & 52
       \\  
       \bf Fast AT~\cite{wong2020fast} & ResNet50 & 61.0 & 43.5 & 12
       \\ 
       \bf ATTA~\cite{zheng2020efficient} & ResNet50 & 60.7 & 44.5 & -
       \\ 
       \hline
       \bf AFT & ResNet50 &  69.5 & 43.0 & 32
       \\
       \bf AFT & ResNet50-SWSL &  \bf 74.5 & \bf 50.5 & 32
       \label{tab:ImageNet}
    %   \vspace{0.7cm}
\end{tabular}
\vspace{-0.5 cm}
\end{table}
\vspace{-0.3 cm}
\section{Conclusion}
\vspace{-0.2 cm}
Here, we further illustrated  the severe overfitting  issue with adversarial training and we argued why this phenomena takes place. Motivated by the finding and experimental results, we proposed simple yet effective fine-tuning approach to improve the robustness of deep neural network models against adversarial attacks without sacrificing the generalization of the model on natural data samples. The proposed fine-tuning framework can reduce the training run-time by $10\times$ while outperforms state-of-the-art algorithms in adversarial training. One important benefit of the proposed method is that it can be easily applied on any pre-trained model without requiring to trained the model from scratch. This is very crucial when the model is trained via customized training frameworks which it is impracticable to train the model again while it is important to improve the robustness of that against adversarial attacks. 

% \section{method}

% \begin{itemize}
% \item Easy samples to hard samples learning
% \item Curriculum learning have shown the advantage of that 
% \item faster convergence
% \item The space of parameters of the target model when is trained on both clean and perturbed samples is somewhere between the space of model trained only on clean data and a model trained only on perturbed samples.
% \item A model trained on both clean and perturbed samples is biased toward perturbed samples.

% \end{itemize}

%--------------------------------------------------------------
{\small
\bibliographystyle{ieee_fullname}
\bibliography{Main}

\begin{thebibliography}{10}\itemsep=-1pt

\bibitem{alayrac2019labels}
Jean-Baptiste Alayrac, Jonathan Uesato, Po-Sen Huang, Alhussein Fawzi, Robert
  Stanforth, and Pushmeet Kohli.
\newblock Are labels required for improving adversarial robustness?
\newblock In {\em Advances in Neural Information Processing Systems}, pages
  12214--12223, 2019.

\bibitem{athalye2018obfuscated}
Anish Athalye, Nicholas Carlini, and David Wagner.
\newblock Obfuscated gradients give a false sense of security: Circumventing
  defenses to adversarial examples.
\newblock {\em arXiv preprint arXiv:1802.00420}, 2018.

\bibitem{carlini2017towards}
Nicholas Carlini and David Wagner.
\newblock Towards evaluating the robustness of neural networks.
\newblock In {\em 2017 ieee symposium on security and privacy (sp)}, pages
  39--57. IEEE, 2017.

\bibitem{carmon2019unlabeled}
Yair Carmon, Aditi Raghunathan, Ludwig Schmidt, John~C Duchi, and Percy~S
  Liang.
\newblock Unlabeled data improves adversarial robustness.
\newblock In {\em Advances in Neural Information Processing Systems}, pages
  11192--11203, 2019.

\bibitem{cohen2019certified}
Jeremy~M Cohen, Elan Rosenfeld, and J~Zico Kolter.
\newblock Certified adversarial robustness via randomized smoothing.
\newblock {\em arXiv preprint arXiv:1902.02918}, 2019.

\bibitem{goodfellow2014explaining}
Ian~J Goodfellow, Jonathon Shlens, and Christian Szegedy.
\newblock Explaining and harnessing adversarial examples.
\newblock {\em arXiv preprint arXiv:1412.6572}, 2014.

\bibitem{he2016deep}
Kaiming He, Xiangyu Zhang, Shaoqing Ren, and Jian Sun.
\newblock Deep residual learning for image recognition.
\newblock In {\em Proceedings of the IEEE conference on computer vision and
  pattern recognition}, pages 770--778, 2016.

\bibitem{he2019parametric}
Zhezhi He, Adnan~Siraj Rakin, and Deliang Fan.
\newblock Parametric noise injection: Trainable randomness to improve deep
  neural network robustness against adversarial attack.
\newblock In {\em Proceedings of the IEEE Conference on Computer Vision and
  Pattern Recognition}, pages 588--597, 2019.

\bibitem{hendrycks2019using}
Dan Hendrycks, Mantas Mazeika, Saurav Kadavath, and Dawn Song.
\newblock Using self-supervised learning can improve model robustness and
  uncertainty.
\newblock In {\em Advances in Neural Information Processing Systems}, pages
  15663--15674, 2019.

\bibitem{ilyas2019adversarial}
Andrew Ilyas, Shibani Santurkar, Dimitris Tsipras, Logan Engstrom, Brandon
  Tran, and Aleksander Madry.
\newblock Adversarial examples are not bugs, they are features.
\newblock In {\em Advances in Neural Information Processing Systems}, pages
  125--136, 2019.

\bibitem{jeddi2020learn2perturb}
Ahmadreza Jeddi, Mohammad~Javad Shafiee, Michelle Karg, Christian
  Scharfenberger, and Alexander Wong.
\newblock Learn2perturb: an end-to-end feature perturbation learning to improve
  adversarial robustness.
\newblock In {\em Proceedings of the IEEE/CVF Conference on Computer Vision and
  Pattern Recognition}, pages 1241--1250, 2020.

\bibitem{kannan2018adversarial}
Harini Kannan, Alexey Kurakin, and Ian Goodfellow.
\newblock Adversarial logit pairing.
\newblock {\em arXiv preprint arXiv:1803.06373}, 2018.

\bibitem{kingma2014adam}
Diederik~P Kingma and Jimmy Ba.
\newblock Adam: A method for stochastic optimization.
\newblock {\em arXiv preprint arXiv:1412.6980}, 2014.

\bibitem{krizhevsky2009learning}
Alex Krizhevsky, Geoffrey Hinton, et~al.
\newblock Learning multiple layers of features from tiny images.
\newblock 2009.

\bibitem{kurakin2016adversarial}
Alexey Kurakin, Ian Goodfellow, and Samy Bengio.
\newblock Adversarial examples in the physical world.
\newblock {\em arXiv preprint arXiv:1607.02533}, 2016.

\bibitem{li2019certified}
Bai Li, Changyou Chen, Wenlin Wang, and Lawrence Carin.
\newblock Certified adversarial robustness with additive noise.
\newblock In {\em Advances in Neural Information Processing Systems}, pages
  9464--9474, 2019.

\bibitem{liu2018adv}
Xuanqing Liu, Yao Li, Chongruo Wu, and Cho-Jui Hsieh.
\newblock Adv-bnn: Improved adversarial defense through robust bayesian neural
  network.
\newblock {\em arXiv preprint arXiv:1810.01279}, 2018.

\bibitem{ma2018characterizing}
Xingjun Ma, Bo Li, Yisen Wang, Sarah~M Erfani, Sudanthi Wijewickrema, Grant
  Schoenebeck, Dawn Song, Michael~E Houle, and James Bailey.
\newblock Characterizing adversarial subspaces using local intrinsic
  dimensionality.
\newblock {\em arXiv preprint arXiv:1801.02613}, 2018.

\bibitem{madry2017towards}
Aleksander Madry, Aleksandar Makelov, Ludwig Schmidt, Dimitris Tsipras, and
  Adrian Vladu.
\newblock Towards deep learning models resistant to adversarial attacks.
\newblock {\em arXiv preprint arXiv:1706.06083}, 2017.

\bibitem{mahajan2018exploring}
Dhruv Mahajan, Ross Girshick, Vignesh Ramanathan, Kaiming He, Manohar Paluri,
  Yixuan Li, Ashwin Bharambe, and Laurens van~der Maaten.
\newblock Exploring the limits of weakly supervised pretraining.
\newblock In {\em Proceedings of the European Conference on Computer Vision
  (ECCV)}, pages 181--196, 2018.

\bibitem{papernot2016distillation}
Nicolas Papernot, Patrick McDaniel, Xi Wu, Somesh Jha, and Ananthram Swami.
\newblock Distillation as a defense to adversarial perturbations against deep
  neural networks.
\newblock In {\em 2016 IEEE Symposium on Security and Privacy (SP)}, pages
  582--597. IEEE, 2016.

\bibitem{schmidt2018adversarially}
Ludwig Schmidt, Shibani Santurkar, Dimitris Tsipras, Kunal Talwar, and
  Aleksander Madry.
\newblock Adversarially robust generalization requires more data.
\newblock In {\em Advances in Neural Information Processing Systems}, pages
  5014--5026, 2018.

\bibitem{shafahi2019adversarial}
Ali Shafahi, Mahyar Najibi, Mohammad~Amin Ghiasi, Zheng Xu, John Dickerson,
  Christoph Studer, Larry~S Davis, Gavin Taylor, and Tom Goldstein.
\newblock Adversarial training for free!
\newblock In {\em Advances in Neural Information Processing Systems}, pages
  3358--3369, 2019.

\bibitem{stutz2019disentangling}
David Stutz, Matthias Hein, and Bernt Schiele.
\newblock Disentangling adversarial robustness and generalization.
\newblock In {\em Proceedings of the IEEE Conference on Computer Vision and
  Pattern Recognition}, pages 6976--6987, 2019.

\bibitem{sun2019towards}
Ke Sun, Zhanxing Zhu, and Zhouchen Lin.
\newblock Towards understanding adversarial examples systematically: Exploring
  data size, task and model factors.
\newblock {\em arXiv preprint arXiv:1902.11019}, 2019.

\bibitem{szegedy2013intriguing}
Christian Szegedy, Wojciech Zaremba, Ilya Sutskever, Joan Bruna, Dumitru Erhan,
  Ian Goodfellow, and Rob Fergus.
\newblock Intriguing properties of neural networks.
\newblock {\em arXiv preprint arXiv:1312.6199}, 2013.

\bibitem{tanay2016boundary}
Thomas Tanay and Lewis Griffin.
\newblock A boundary tilting persepective on the phenomenon of adversarial
  examples.
\newblock {\em arXiv preprint arXiv:1608.07690}, 2016.

\bibitem{thomee2016yfcc100m}
Bart Thomee, David~A Shamma, Gerald Friedland, Benjamin Elizalde, Karl Ni,
  Douglas Poland, Damian Borth, and Li-Jia Li.
\newblock Yfcc100m: The new data in multimedia research.
\newblock {\em Communications of the ACM}, 59(2):64--73, 2016.

\bibitem{torralba200880}
Antonio Torralba, Rob Fergus, and William~T Freeman.
\newblock 80 million tiny images: A large data set for nonparametric object and
  scene recognition.
\newblock {\em IEEE transactions on pattern analysis and machine intelligence},
  30(11):1958--1970, 2008.

\bibitem{tramer2017ensemble}
Florian Tram{\`e}r, Alexey Kurakin, Nicolas Papernot, Ian Goodfellow, Dan
  Boneh, and Patrick McDaniel.
\newblock Ensemble adversarial training: Attacks and defenses.
\newblock {\em arXiv preprint arXiv:1705.07204}, 2017.

\bibitem{tsipras2018robustness}
Dimitris Tsipras, Shibani Santurkar, Logan Engstrom, Alexander Turner, and
  Aleksander Madry.
\newblock Robustness may be at odds with accuracy.
\newblock {\em arXiv preprint arXiv:1805.12152}, 2018.

\bibitem{wong2020fast}
Eric Wong, Leslie Rice, and J~Zico Kolter.
\newblock Fast is better than free: Revisiting adversarial training.
\newblock {\em arXiv preprint arXiv:2001.03994}, 2020.

\bibitem{xie2019feature}
Cihang Xie, Yuxin Wu, Laurens van~der Maaten, Alan~L Yuille, and Kaiming He.
\newblock Feature denoising for improving adversarial robustness.
\newblock In {\em Proceedings of the IEEE Conference on Computer Vision and
  Pattern Recognition}, pages 501--509, 2019.

\bibitem{yalniz2019billion}
I~Zeki Yalniz, Herv{\'e} J{\'e}gou, Kan Chen, Manohar Paluri, and Dhruv
  Mahajan.
\newblock Billion-scale semi-supervised learning for image classification.
\newblock {\em arXiv preprint arXiv:1905.00546}, 2019.

\bibitem{yuan2019adversarial}
Xiaoyong Yuan, Pan He, Qile Zhu, and Xiaolin Li.
\newblock Adversarial examples: Attacks and defenses for deep learning.
\newblock {\em IEEE transactions on neural networks and learning systems},
  30(9):2805--2824, 2019.

\bibitem{zagoruyko2016wide}
Sergey Zagoruyko and Nikos Komodakis.
\newblock Wide residual networks.
\newblock {\em arXiv preprint arXiv:1605.07146}, 2016.

\bibitem{zhang2019you}
Dinghuai Zhang, Tianyuan Zhang, Yiping Lu, Zhanxing Zhu, and Bin Dong.
\newblock You only propagate once: Accelerating adversarial training via
  maximal principle.
\newblock In {\em Advances in Neural Information Processing Systems}, pages
  227--238, 2019.

\bibitem{zhang2019defense}
Haichao Zhang and Jianyu Wang.
\newblock Defense against adversarial attacks using feature scattering-based
  adversarial training.
\newblock In {\em Advances in Neural Information Processing Systems}, pages
  1831--1841, 2019.

\bibitem{zhang2019theoretically}
Hongyang Zhang, Yaodong Yu, Jiantao Jiao, Eric~P Xing, Laurent~El Ghaoui, and
  Michael~I Jordan.
\newblock Theoretically principled trade-off between robustness and accuracy.
\newblock {\em arXiv preprint arXiv:1901.08573}, 2019.

\bibitem{zheng2020efficient}
Haizhong Zheng, Ziqi Zhang, Juncheng Gu, Honglak Lee, and Atul Prakash.
\newblock Efficient adversarial training with transferable adversarial
  examples.
\newblock In {\em Proceedings of the IEEE/CVF Conference on Computer Vision and
  Pattern Recognition}, pages 1181--1190, 2020.

\end{thebibliography}
}

\end{document}